\documentclass[a4paper,10pt]{article}

\usepackage{epsfig}
\usepackage{latexsym}
\usepackage{layout}
\usepackage{graphicx}

\title{\textbf{Coordinated exploration of labyrinthine environments with application to the ``pursuit evasion'' problem}}
\author{Damien Pellier \\
Damien.Pellier@imag.fr
\and
Humbert Fiorino\\
Humbert.Fiorino@imag.fr}

\date{Laboratoire Leibniz (CNRS - INPG - IMAG)\\
46, Avenue Felix Viallet, F-38031 Grenoble Cedex \\}

\begin{document}

\maketitle

\abstract{This paper introduces a multirobot cooperation approach to solve the ``pursuit evasion''
problem for mobile robots that have omnidirectional vision sensors. The main characteristic of this approach is to implement a real cooperation between robots based on knowledge sharing and makes them work as a team.
A complete algorithm for computing a motion strategy of robots is also presented. This algorithm is based on searching critical points in the environment. Finally, the deliberation protocol which distributes the exploration task among the team and takes the best possible outcome from the robots resources is presented.\\}

{\bf keywords:} coordinated exploration\ ---\ multirobot cooperation\ ---\ ``pursuit evasion'' problem

\section{Introduction}
The domain of cooperative robotics is acquiring a prominent interest in many applications such as drones or Unmanned Autonomous Vehicles formations \cite{Vidal-Rashid-Sharp-Shakernia-Kim-Satry-01}, mobile robots carrying out transportation tasks \cite{Alami-Fleury-Herrb-Ingrand-Robert-98}, exploring the environment \cite{Lang-Chee-98}, interacting with people \cite{Drogoul-Picault-00} etc. To that end, many different approaches have been proposed ranging from reactive behaviors \cite{Balch-Arkin-98} to deliberative protocols \cite{Alami-Fleury-Herrb-Ingrand-Robert-98}.\\

In this paper, we consider the following problem: a group of robots have to explore a labyrinthine environment and find an intruder if need be. Such a problem has been already studied in mainly two different aspects.\\

On the one hand, the ``prey predators'' problem has been proposed for the first time by \cite{Benda-Jagannathan-Dodhiawala-86}: the prey and the predators, i.e. the {\it agents}, share a common environment which is represented by a mere grid. The hunt is simulated and the agents that are autonomous processes can move horizontally or vertically all the time. Quite obviously, the {\it goal} of the prey is to escape as long as possible whereas the predators have to capture the prey as fast as possible. Therefore, the ``prey predators'' problem is an interesting testbed for competing agents and coordination protocols. Indeed, because each agent does not perceive all the environment, and can have incomplete or inconsistent knowledge about the other agents, many challenging issues have to be tackled: which information has to be communicated? to whom? when? how can the predators constrain the prey movements? how can they elaborate a common strategy and behave as a {\it team}? otherwise, can the predators exclusively rely on reactive behaviors? which is the best approach in terms of performances, implementation, communication hazards (communication bottleneck, deadlocks etc.) ? Unfortunately, the notion of environment is not considered in those experiments.\\

\begin{figure}
\begin{minipage}[b]{.46\linewidth}
\centering\includegraphics[scale=0.33]{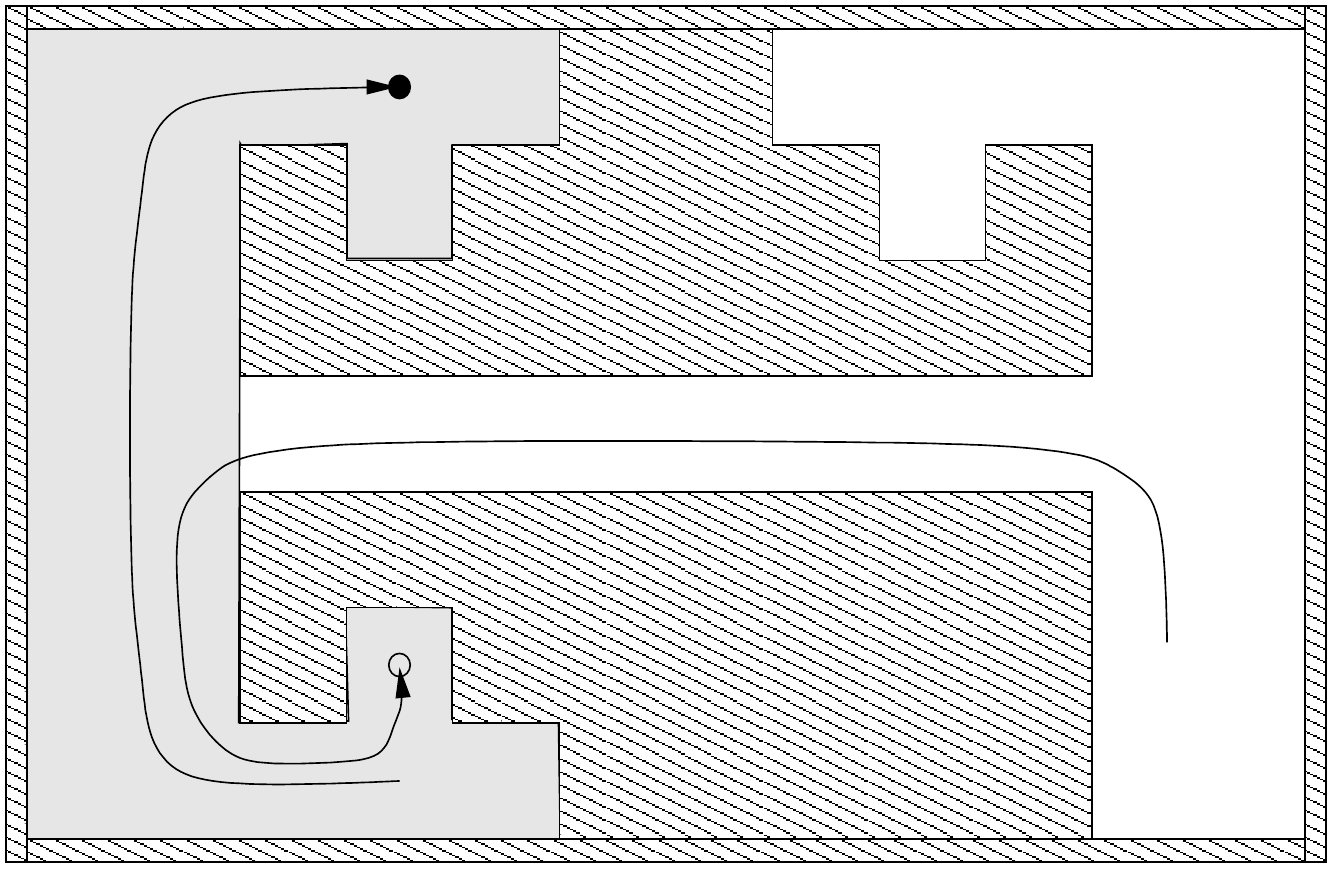}
\caption{An intruder can move under cover from a hiding place to places already explored by the pursuer.}
\label{Figure:Evader}
\end{minipage} \hfill
\begin{minipage}[b]{.46\linewidth}
\centering\includegraphics[scale=0.4]{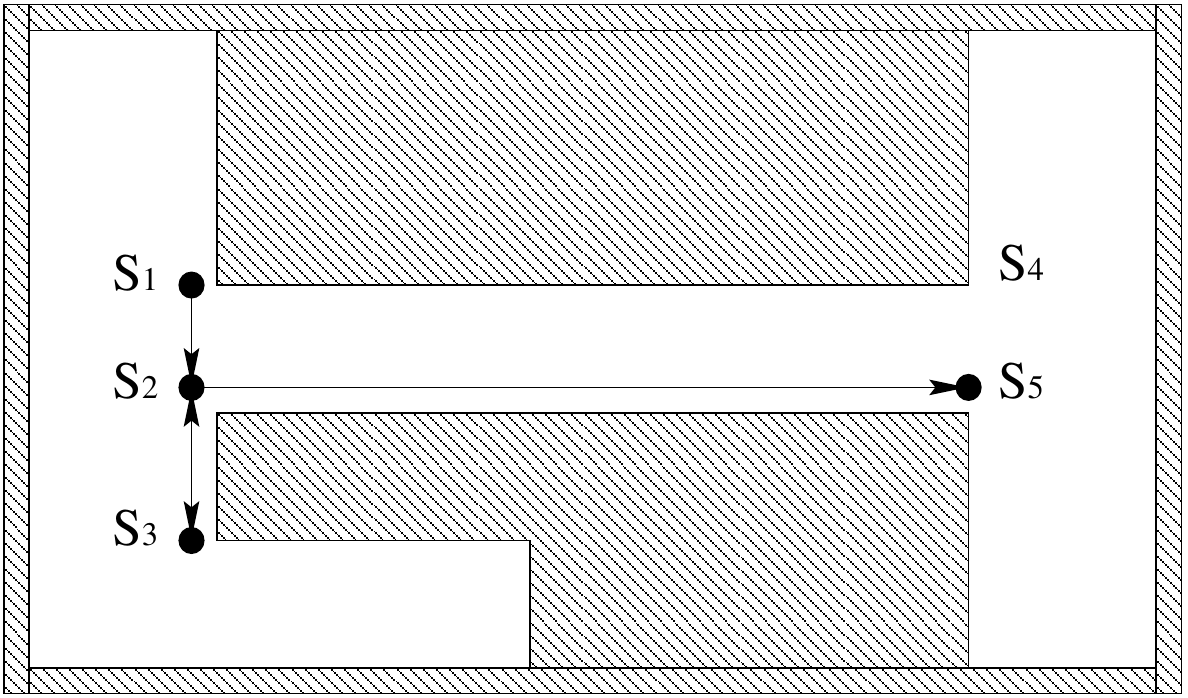}
\caption[2D Euclidean space including polygonal obstacles]{The {\textbullet} represents a pursuer and its trajectory $\langle S_{1},S_{2},S_{3},S_{2},S_{5}\rangle$ in a 2D Euclidean space including polygonal obstacles.}
\label{Figure:Simple-example}
\end{minipage}
\end{figure}

On the other hand, the ``pursuit evasion'' problem is based on an environment made up of many obstacles. It is postulated that an intruder may be hidden within this environment, and a pursuer must flush it out of its hiding place. Therefore, a solution is a path ensuring that whatever movement is realized by the intruder, finally it will be uncovered by the pursuer. This problem has been tackled in many ways such as game theory \cite{Levy-Rosenschein-92}, graph theory \cite{Bienstock-Seymour-91,Lapaugh-93,Makedon-SudBorough-83,Megiddo-Hakini-Garey-Johnson-Papadimitriou-88,Monien-Subborough-88,Parsons-76} etc. As far as robotics are concerned, the ``pursuit evasion'' problem has been introduced by \cite{Suzuki-Yamashita-92}. Since then, several works have been undertaken: in \cite{Lee-Park-Chwa-00}, robots fitted with one detection beam must keep watch on a grid environment with one exit; in \cite{Icking-Klein-91,Heffernan-93}, two keepers move along the borders of a labyrinthine environment without isolated obstacles (i.e., that are not bound to a border) and have to keep in touch constantly with their sensors. The ``pursuit evasion'' problem has also been looked at more generally in \cite{Crass-Suzuki-Yamashita-95,LaValle-Hinrichnsen-99,LaValle-Lin-Guibas-Latombe-Motwani-97,Suzuki-Yamashita-92}: this time, there is no restriction on the environment (2D polygonal or curved environment) and the pursuers are fitted with omnidirectional detection beams. However, either the environment can be explored by only one pursuer and then the algorithm provides the path ensuring that the intruder (its velocity can be arbitrary high) will be uncovered, or, because of the environment topology, several pursuers are needed and then the algorithm provides the path that has to be followed by {\it one} pursuer completed by {\it observation posts} for the other robots. This is due to the intruder's ability to move under cover from a hiding place to places already explored by the pursuer (cf. the intruder represented by a blank dot, figure \ref{Figure:Evader}).

In order to avoid the drawbacks of the former approaches, we present a cooperation protocol that allows the pursuers to coordinate their exploration and jointly look after the intruder. Therefore, they operate like a team and the number of pursuers is ``minimized''. In section \ref{Problem}, the pursuit evasion problem is defined. In section \ref{OnePursuer}, a complete algorithm for one pursuer is presented. In section \ref{Cooperation}, we introduces the deliberation protocol and the section \ref{Simulation} shows some computed examples.

\section{The pursuit evasion problem}
\label{Problem}
In this paper, we assume that the robots manoeuvre within a 2D polygonal environment (cf. figure \ref{Figure:Simple-example}); their vision is omnidirectional and, at first, we assume that the whole environment is known.\\

The pursuers and the intruder are represented by points in the 2D Euclidean space. Let $F$ be the free space (obviously, the pursuers and the intruder belong to $F$). Let $e(t) \in F$ be the position of the intruder at time $t \geq 0$. We suppose that $e~: [0,\infty[ \rightarrow F$ is a continuous function and that the intruder is able to move as fast as it wants: its initial position $e(0)$ and its trajectory $e$ are not known by the pursuers. Let $\gamma^{i}(t) $ be the position of the i$th$ pursuer at time $t \geq 0$. $\gamma^{i}: [0,\infty[ \rightarrow F$ is the continuous trajectory of the i$th$ pursuer. Let $\gamma$ be the trajectories set of the $N$ pursuers~: $\gamma = \{\gamma^{1},\ldots,\gamma^{N}\}$. For all point $q \in F$, let $V(q)$ be the set of all visible points from $q$ in $F$ (i.e., all the segments joining $q$ to a point of $V(q)$ strictly belong to $F$). The trajectory or ``watch'' $\gamma$ is a solution if, for all continuous function $e: [0,\infty[ \rightarrow F$, $t \in [0,\infty[$ such that $e(t) \in V(\gamma^{i}(t))$ (with $i \in \{1,\ldots, N\}$) exists. This implies that the intruder cannot escape; at a given time, necessarily, it will be uncovered.\\

Let $H(F)$ be the minimum number of pursuers such that there is a trajectory ensuring that the intruder will be uncovered. According to \cite{Guibas-Latombe-LaValle-Lin-Motwani-96}, in the worst case $H(F) = O(h + log\ n)$ in an environment made up of isolated obstacles where $n$ is the number of edges and $h$ the  number of holes, and $H(F) = O(log\ n)$ otherwise. Therefore, two different problems have to be distinguished:
\begin{enumerate}
\item $H(F)$ calculation;
\item For a given $F$, find the trajectory $\gamma$ implementing at the most $H(F)$ pursuers.
\end{enumerate}

\section{Complete algorithm for one pursuer}
\label{OnePursuer}
\subsection{General principle}
This algorithm is inspired by \cite{LaValle-Lin-Guibas-Latombe-Motwani-97}. One of the main difficulties that have to be dealt in the pursuit evasion problem is that the intruder can move under cover from a hiding place to places already explored by the pursuer (cf. figure \ref{Figure:Evader}). Consequently, the pursuer trajectory must all the time consider all the possible moves of the intruder. Thus, all the areas of $F$ that can include the intruder are called ``contaminated'' and ``decontaminated'' otherwise. In other words, the purpose of the pursuer is to entirely decontaminate $F$: the environment has to be divided up into convex areas where the intruder can hide.\\

\begin{figure}
\begin{center}
\centering\includegraphics[scale=0.55]{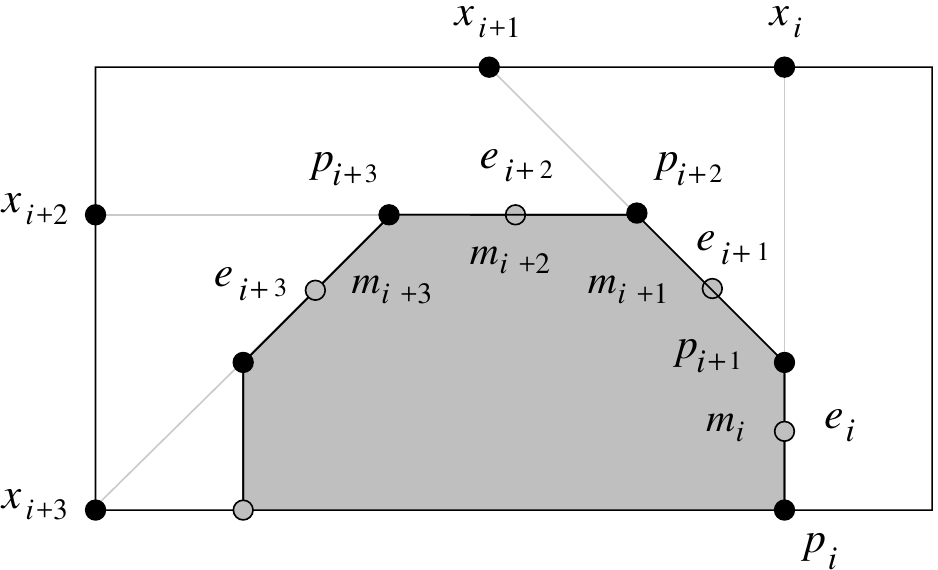}
\caption{Identification of the critical points.}
\label{Figure:Critical-points}
\end{center}
\end{figure}

Let $P$ be an obstacle and $\partial P$ the set of its edges. Consider a vertex $p_{i} \in \partial P$. This vertex is called a \textit{critical point} if the internal angle formed by the edges $e_{i-1}$ and $e_{i}$ is inferior to $\pi$ (cf. figure \ref{Figure:Critical-points}); $CP$ is the set of all the critical points. Intuitively, if we could put a robot with an omnidirectional vision on each critical point, there would have no place where the intruder could hide in $F$. In this paper, we assume that the pursuers move from one critical point to another one in order to find the intruder.\\

Let $m_{i}$ be the middle of the edge $e_{i}$. If we draw the segment joining $m_{i}$ through $p_{i}$ until $x_{i}$, the intersection of this segment with the environment border, a convex area is created where the intruder cannot hide when the pursuer moves along $[p_{i}, p_{i+1}[$. However, the intruder can hide outside the ``frontier'' $[m_{i},x_{i}]$. Therefore, this procedure provides a division of $F$ into convex cells that have to be visited by the pursuer.\\

Let $y_{i}$ be the position of the $i^{th}$ pursuer, $0 \leq i \leq N$. A critical point $p_{j} \in CP$ is ``decontaminated'' or {\it cleared} if $p_{j}$ is visible from at least one pursuer placed on $y_{i}$ and its adjacent edges $p_{j}$ are totally visible from $y_{i}$.\\

The state of the environment is represented by the set of the critical points visible from this point, and the labels ``cleared'' or ``contaminated'' are associated with these points. The states set of the points $y_{i}$ where the pursuer (resp. the team of pursuers) is situated, provides the global state of the environment at time $t \geq 0$ of the trajectory. The problem that the pursuer (resp. the team of pursuers) must solve comes down to find a trajectory $\gamma$ that goes through $CP$ and ends in a final state such as all the points in $CP$ are cleared.

\subsection{Construction of the trajectory}

This algorithm calculates the optimal trajectory in terms of covered distance. The starting point for the exploration is arbitrary chosen in $CP$. The algorithm is made up of four steps:
\begin{enumerate}
\item First of all, the critical points have to be found (cf. figure \ref{Figure:Simple-example}: the critical points are numbered from 1 to 5). The identification procedure based on angles calculation has a linear cost that depends on the number of the critical points in the environment (cf. figure \ref{Figure:Critical-points});
\item The algorithm constructs the \textit{visibility graph} $G_{v}$ which nodes are the critical points identified beforehand. From a given critical point, the edges joining all the other critical points visible from that point are created (this graph is necessarily connected if the environment is ``connected'' i.e. for all couple of points from the free space, it exists a path joining them): thus $G_{v} = (X,E)$ with $X = CP$ and $E = \{e_{1}, \ldots, e_{m}\}$; each edge is labeled with the distance between both vertices; $G_{v}$ represents all the possible moves from a given critical point;
\item In that step, the algorithm constructs the \textit{surveillance graph} $G_{s}$ of $F$ from $G_{v}$. The initial state $E_{initial}$ of $G_{s}$ is calculated according to the starting point  $p_{input}$ passed to the algorithm. $E_{initial}$ is made up of the critical points set associated with their corresponding surveillance states (``cleared'' or ``contaminated''). For each critical point $p_{i}$ of the considered node, the ``cleared'' label is given if and only if the vertex $p_{i}$ is visible from $p_{input}$ and the adjacent segments of $p_{i}$ are entirely visible from $p_{input}$. For each possible transition $T = \{t_{1}, \ldots, t_{n}\}$ (where $n$ is the number of critical points visible from $p_{input}$), a new node $E_{i+1}$ is generated applying this procedure. However, when generating these new nodes, it must be verified that the formerly cleared critical points remain cleared in the following node. Therefore, the following condition must be assessed: for all cleared critical points in the current node, there do not exist a path from a contaminated point to one of these points; moreover, if that path exists, the ``threatened'' points remain visible during the transition $t_{i}$. If this condition is verified then the points remain cleared (otherwise they become contaminated). The transition is labeled with the distance previously calculated into $G_{v}$. This step is applied recursively on all new nodes. Thus, because all possible transitions are explored, it is guaranteed that, if the trajectory for one pursuer exists, it will be found;

\item In the last step, the algorithm determines the optimal trajectory allowing the decontamination of $F$: after the identification of all the final states in $G_{s}$ (i.e. all the critical points of a given node are labeled cleared), for each final state, the Dijsktra algorithm is applied in order to find out the optimal trajectory joining each final vertex to the starting point.

\begin{figure}[htbp]
\begin{center}
\centering\includegraphics[scale=0.45]{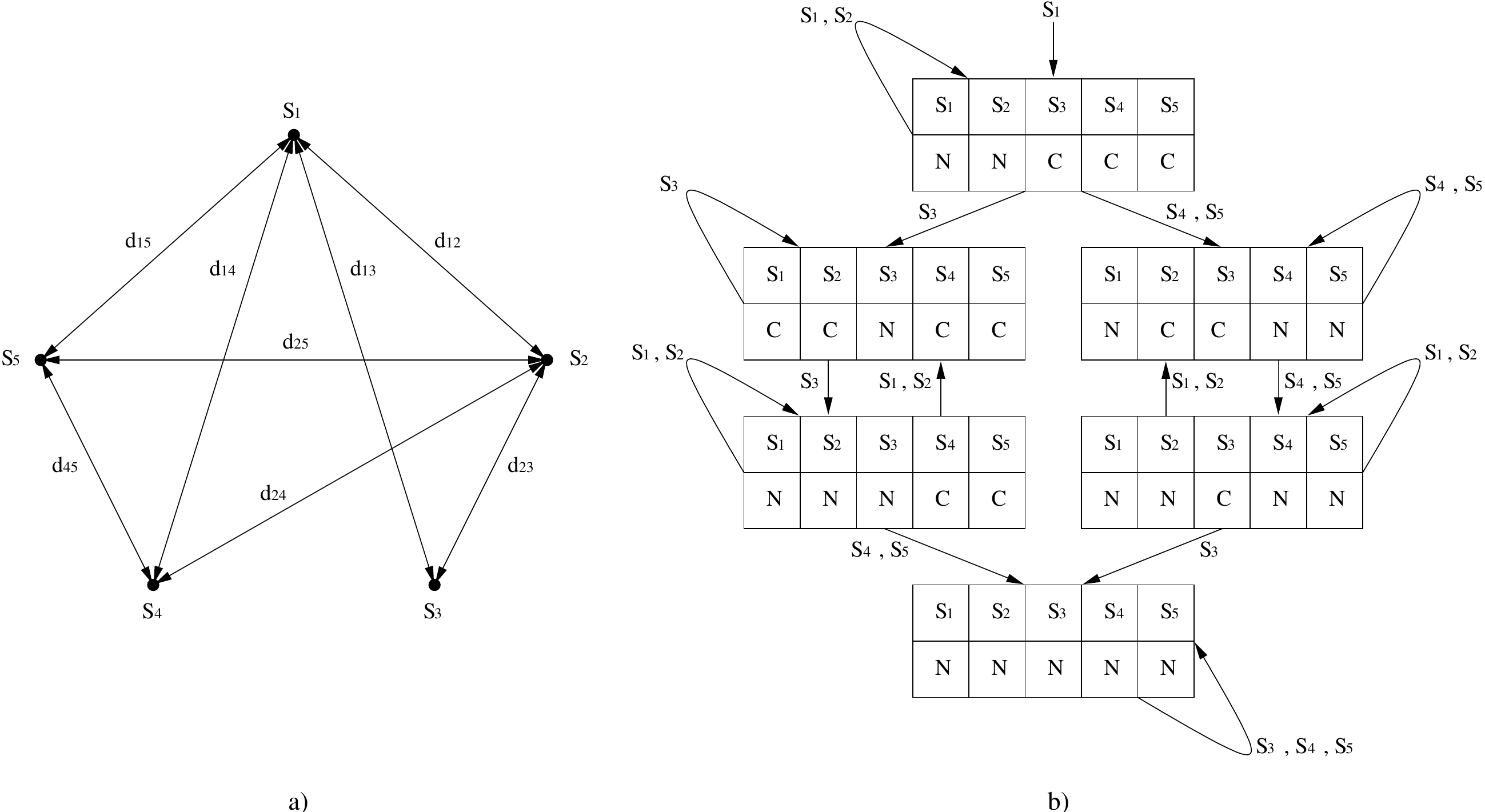}
\caption[a) visibility graph b) surveillance graph]{a) shows the visibility graph and b) the surveillance graph corresponding to figure \ref{Figure:Simple-example}: ``N'' means that the vertex is cleared; ``C'' means contaminated.}
\label{Figure:Monitoring-graph}
\end{center}
\end{figure}
\end{enumerate}

\section{Cooperation implementation}
\label{Cooperation}
In the previous section, an algorithm for one pursuer was described. The graph $G_{s}$ can be built without the guarantee that a final state will be found. In this case, the environment must be explored at least by one more robot and a cooperative approach is proposed. However, this approach does not calculate a trajectory for a given number of pursuers: it calculates a trajectory that keeps the number of robots as close as possible of $H(F)$ by adding additional robots when cooperation fails (the calculation of $H(F)$ according to the environment is a NP-Hard problem).

\subsection{Detection of the ``delegation points''}

When a pursuer do not find a solution to decontaminate the environment, it decides to explore its surveillance graph and prune all the edges that lead to a recontaminated environment. The obtained graph has the property of not allowing the recontaminating moves. This property of the surveillance graph is verified when at least one cleared critical point constituting a node of the surveillance graph remains cleared in the successor node. In that case, the successor node is not removed. We call {\it delegation point} a point $p_{d}$ where all edges do not satisfy this property. The pursuer records this point for a future use. Thus at this point $p_{d}$, the pursuer needs another pursuer in order to help it to proceed to the exploration. The delegation points are represented in the graph by a node where the pursuer cannot move without recontaminating a part of the environment: $p_{d}$ is a sink of the surveillance graph.
Because of the exploration of the graph in depth first, it is possible to find more than one delegation point. For each delegation point, the algorithm constructs a surveillance trajectory from its current position to $p_{d}$ as if $p_{d}$ was a final state (see section \ref{OnePursuer}). The chosen delegation point is the nearest one.

\subsection{Need for assistance}

Whenever a robot reaches a delegation point, the teammates must be able to coordinate themselves in order to split the task of this robot into a set of sub-tasks.  Before asking the others robots for help, the ``stuck'' robot must compute the points of the environment that need to be kept under surveillance so as it can continue its exploration. To this end, the robot at the delegation point tries to divide its visibility graph in order to delegate the control over some parts of the environment: two cut methods are considered. The goal of these methods is to break up the environment into parts that can be watched by independent robots.\\

The first one is called the {\it simple cut} (see figure \ref{Figure:Example-1}). The robot removes the delegation points $p_{d}$ of its visibility graph $G_{v}$. If $G_{v}$ is broken into connected components, the robot obtains a decomposition of the environment in parts that can be explored by independent robots. This division allows to spread and parallelize the environment exploration. As the withdrawal of the delegation points splits the visibility graph, it is quite obvious that the intruder cannot move from different areas (represented by the connected components of $G_{v}$) without passing over this $p_{d}$. Consequently, each sub part of the environment can be explored by one robot or a team of robots independently and simultaneously without any risk of recontamination.\\

But this cut is not always possible (see figure \ref{Figure:Example-2}). In this case, the robot at the delegation point tries to make another cut called {\it complex cut}. The robot removes from the visibility graph $G_{v}$ the delegation point $p_{d}$ but also all the visible points from $p_{d}$. If $G_{v}$ is broken into connected components, the robot knows that it needs a guard robot staying on the delegation point to split the environment into parts that are represented by the different connected components of the visibility graph. Then, each part of the environment can be explored independently. As we said before, leaving a guard robot on the delegation point guarantees that an intruder cannot move from a part of the environment to another one. If this method fails to share the environment into independent components, then we obtain one connected component and the guard robot is still needed on the delegation point: in this case, the exploration cannot be parallelized. But, the environment that remains to be explored (the obtained connected component) is reduced and the division algorithm has to be run recursively on this component.
(see figure \ref{Figure:Example-3}).\\

To summarize, we can distinguish two kinds of assistance tasks that can be required by the stuck robot at a delegation point:
\begin{itemize}
\item Type I: assistance in order to explore independent sub-parts of the environment;
\item Type II: assistance in order to keep some specific points of the environment under surveillance and therefore share the environment into smaller parts.
\end{itemize}

\subsection{Tasks delegation: the deliberation protocol}

\begin{figure}[htbp]
\centering\includegraphics[scale=0.61]{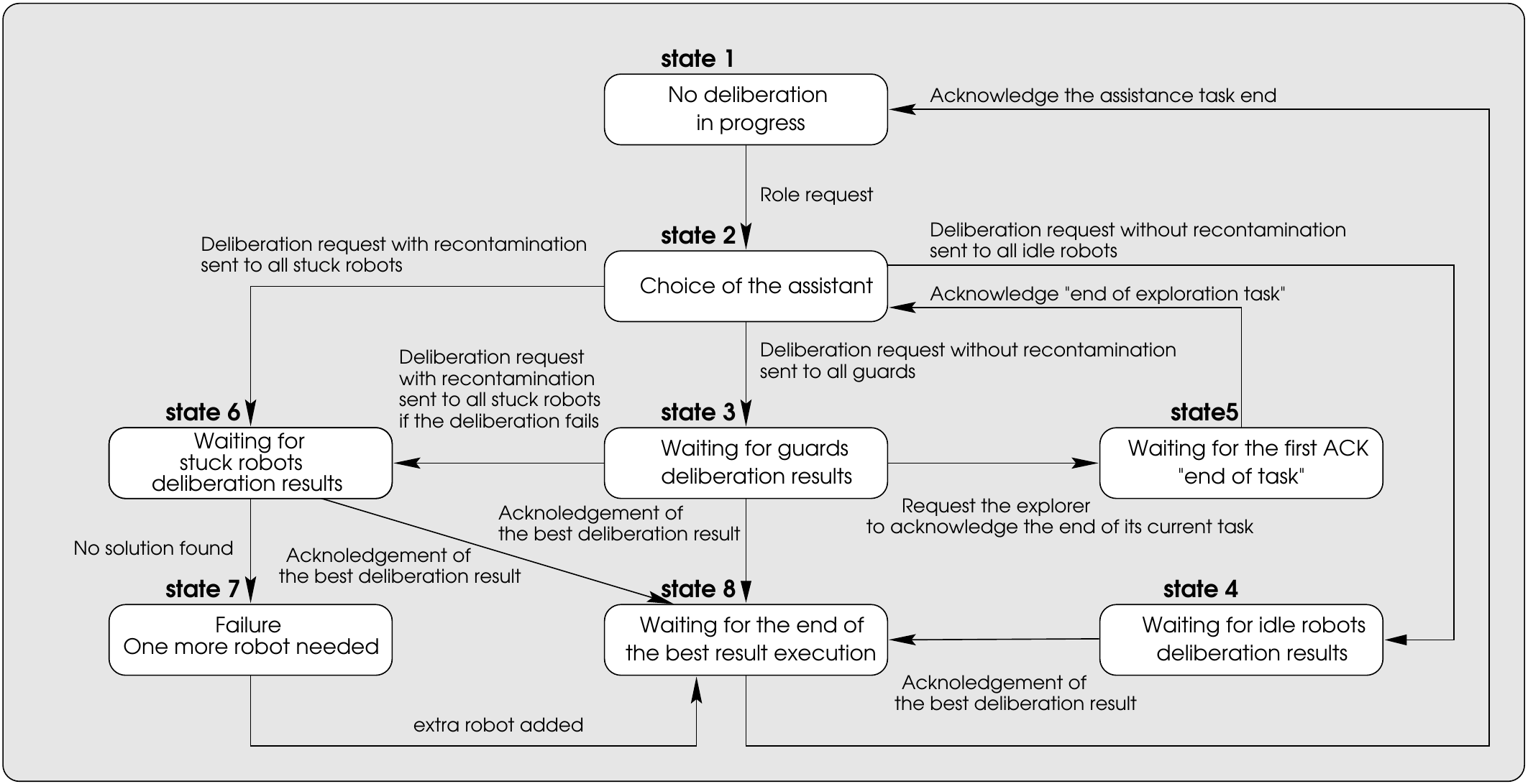}
\caption{Deliberation protocol: side of the assisted robot}
\label{Figure:assisted}
\end{figure}

\begin{figure}[htbp]
\centering\includegraphics[scale=0.7]{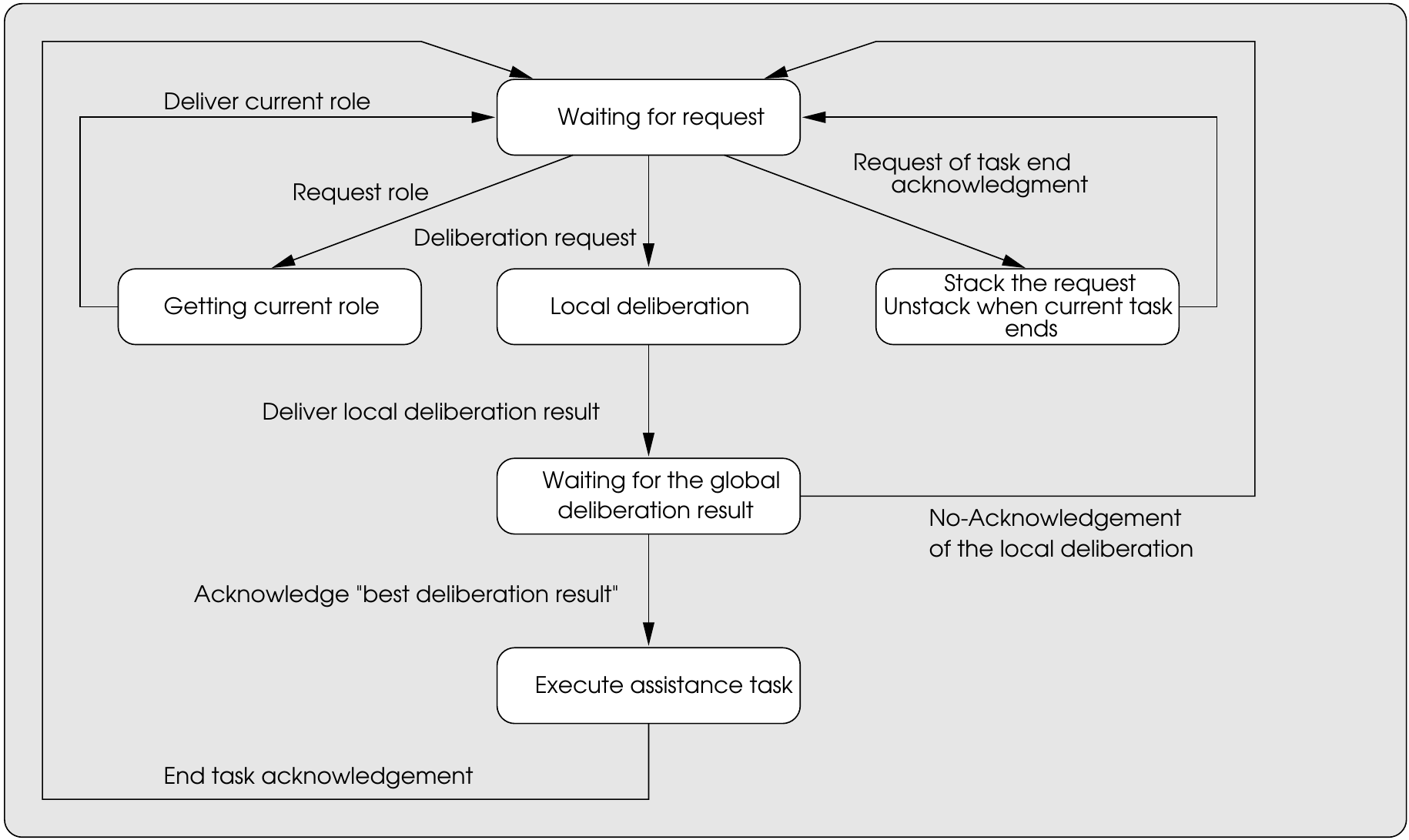}
\caption{Deliberation protocol: side of the assistant robot}
\label{Figure:assistant}
\end{figure}

We saw in the former paragraph how the robot that is stuck at a delegation point determines its needs in terms of surveillance before continuing its exploration. Consequently, the new problem is to find robots ready to provide assistance. The purpose of this section is to introduce a deliberation protocol for the assisted and the assistant robots (see figure \ref{Figure:assisted} and \ref{Figure:assistant} respectively). This protocol is based on the four different ``roles'' that can be played by the robots while exploring the environment (this roles change during the labyrinth exploration):
\begin{itemize}
\item \textit{explorer}: the robot explores an assigned part of the environment;
\item \textit{guard}: the robot is used to share the environment into parts so that an intruder cannot move from one part to another one;
\item \textit{idle robot}: the robot is idle and it can be required for a new task (i.e. it ended previous assigned tasks);
\item \textit{stuck robot}: this robot needs help to continue its exploration. It cannot move without risking a recontamination of previously explored areas.
\end{itemize}

Therefore, the first step of the deliberation protocol is to collect the robots roles (see figure \ref{Figure:assisted}, state 1). A stuck robot runs this protocol in order to estimate the team capabilities  and to request assistance. When this robot receives all the teammates roles, it can assess which robots can help it (see figure \ref{Figure:assisted}, state 2). Three cases are possible:
\begin{itemize}
\item if at least one robot is idle, the stuck robot informs the idle robots of its need for assistance (type I or II) because they can be immediatly reallocated without any risk of recontamination (see figure \ref{Figure:assisted}, state 4);
\item if all robots are busy (i.e, guard or explorer) but there is at least one guard, the stuck robot sends to the guards the visible points it would like to be watched and the global state of the team  (see figure \ref{Figure:assisted}, state 3). To obtain this global state, the stuck robot requests the local state of each teammate (i.e. the current node of their surveillance graph) and merges all these states. With this information each guard is able to compute all the possible moves it can do in order to assist the stuck robot. They must solve two different kind of problems: ensure the division of the environment and answer to the assistance requests. More precisely, two possibilities must be considered:
\begin{itemize}
\item the guard finds that all the visible points from its position are cleared and thus, it becomes a idle robot. Indeed, if this condition is verified, the guard is sure that its surveillance task is finished. It has no reason to watch points that are cleared because the algorithm for on pursuer ensures that no intruder can be located into a cleared part of the environment. Therefore, it can leave its current position and calculate the best path to reach a point from which it can carry out the requested assistance task;

\item The guard finds that at least one visible point from its position is contaminated. It cannot be reallocated to explore a sub-part of the environment. But, in some cases, it can find a path that drives it to a point from which it can carry out the requested assistance task {\it and} fulfill its current guard task;
\end{itemize}
\item if all robots are stuck (see figure \ref{Figure:assisted}, state 6), the last stuck robot (that finds that all its teammate are also stuck) builds the global state and broadcasts this information with its need for assistance. In order to avoid a deadlock, at least one stuck robot must give up its assistance request and its planned exploration task in order to provide assistance to one of its teammate. Thus, the areas it had previously explored must be considered as re-contaminated. Then, this robot becomes idle and provides assistance to the last stuck robot. To that end, it calculates the cost of its assistance in terms of trajectory distance and number of re-contaminated critical points. Furthermore, the protocol must ensure that the team will not enter into an infinite loop. For instance, assume that two area A and B have to be cleared: if clearing the A area implies to contaminate the B area, it must be ensured that clearing B does not imply to re-contaminate A. The infinite loop is avoided by recording information about the previously cleared and re-contaminated areas. If such a loop is detected, then at least one more robot is needed.
\end{itemize}

After this step, each robot delivers its answer to the assistance request (i.e. ``refused'' or ``accepted'' request and, in the former case, the cost of the assistance task) to the stuck robot that chooses the best solution (according to distance and number of recontaminated points criteria). At this step two cases are possible:
\begin{itemize}
\item The deliberation converges toward one or more solutions. The best solution is chosen according to the previously introduced criteria. The chosen robot receives an acknowledgement that informs it to start the assistance task. The other robots receive a no-acknowledgement to indicate that their solutions were not accepted. Finally, when the assistant robot finishes its task, it delivers to the stuck robot an ``end of task'' message. The former stuck robot can resume its exploration task and becomes an explorer;
\item The global deliberation fails. No solution is available:
\begin{itemize}
\item if no solution was found even with recontamination (a stuck robot abandons its exploration task), the team fails and cannot solve the problem with the number of robot involved into the environment. Necessarily, at least one more robot is needed. This robot is added to the team and the deliberation protocol is run anew in order to find the minimal number of robots needed to converge toward a solution (see figure \ref{Figure:assisted}, state 7);
\item if at least one robot is an explorer (its exploration task is in progress but soon or later it will become idle or stuck), potentially it can decontaminate an area and thus allow guards to become idle robots. Thus, the failure is not effective until there is no more explorer. Anyway, they must inform the stuck robot of the end of their current tasks. Indeed, the global state may have evolved favorably. A new round of deliberation is initiated (in the worst case, the number of additional deliberation rounds is equal to the number of explorer found during the first round) and can lead to a solution (see figure \ref{Figure:assisted}, state 5);
\item if the team is only made up of guards and stuck robots and no solution was found by the guards, the last stuck robot broadcasts to the other stuck robots a deliberation request. Thus, the protocol returns to the state 6 previously described.
\end{itemize}
\end{itemize}

\section{Simulation examples}
\label{Simulation}
\begin{figure}
\begin{minipage}[b]{.46\linewidth}
\centering\includegraphics[scale=0.35]{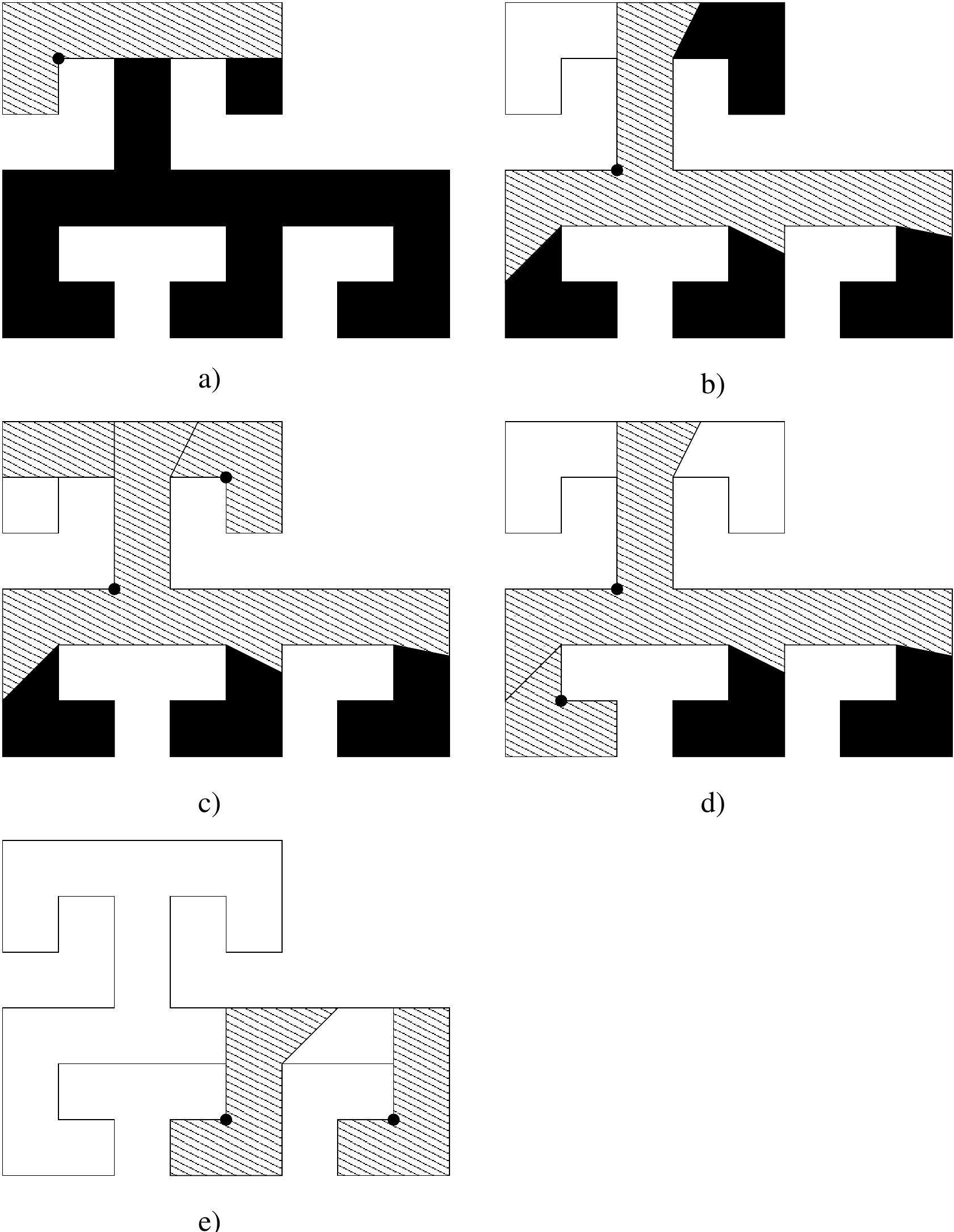}
\caption{Simple cut}
\label{Figure:Example-1}
\end{minipage} \hfill
\begin{minipage}[b]{.46\linewidth}
\centering\includegraphics[scale=0.35]{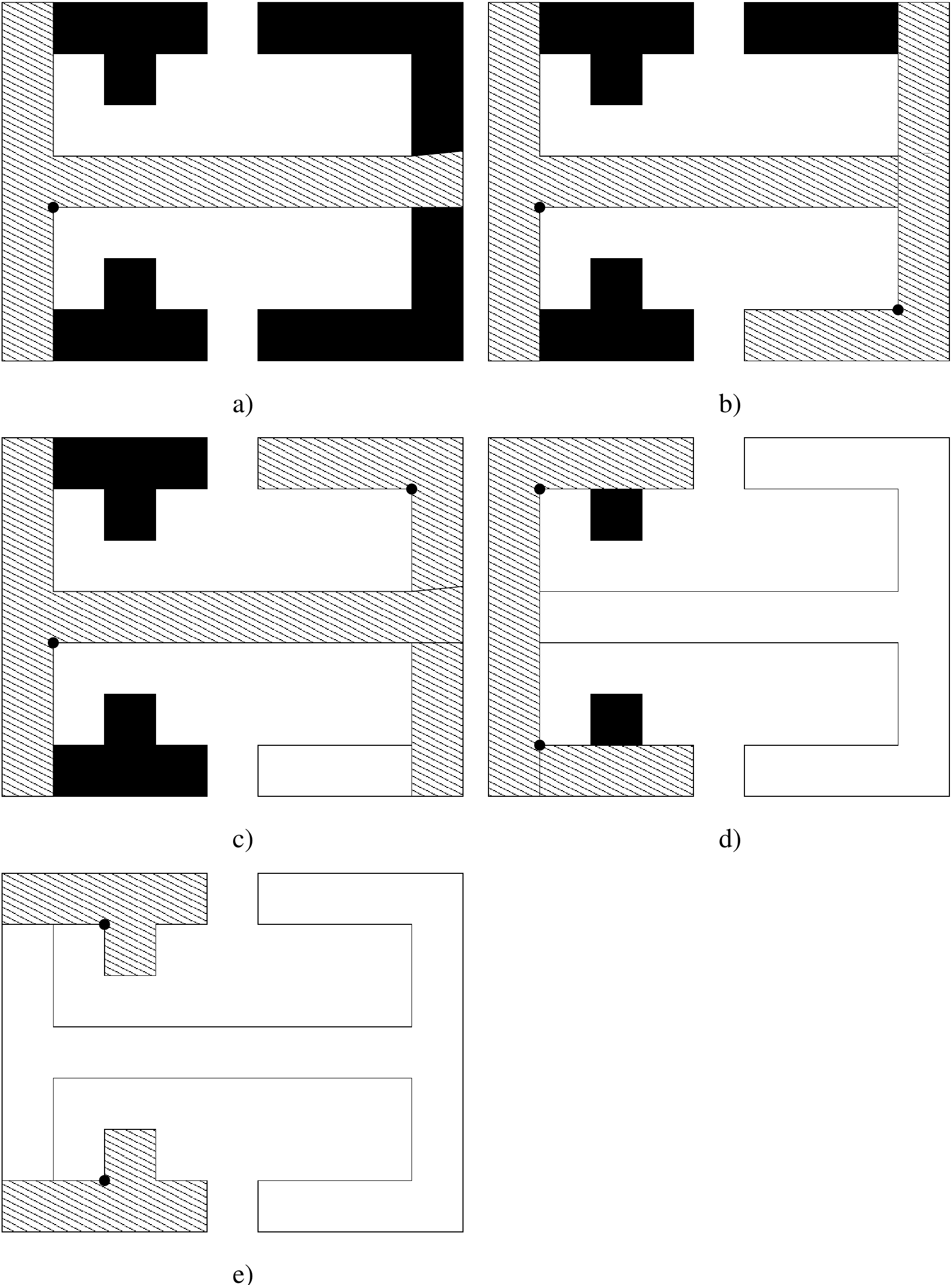}
\caption{Complex cut}
\label{Figure:Example-2}
\end{minipage}
\end{figure}

\begin{figure}[htbp]
\centering\includegraphics[scale=0.5]{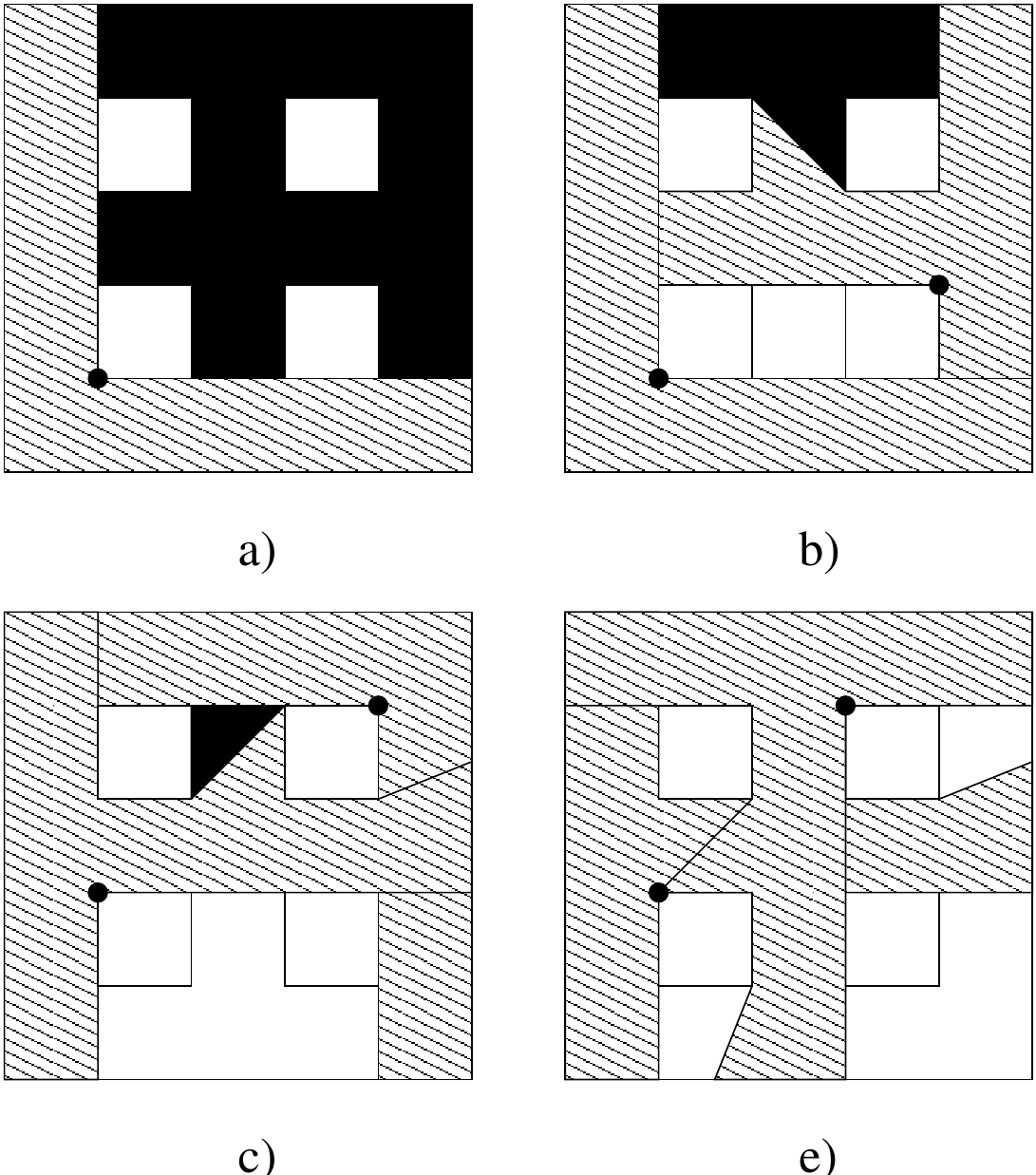}
\caption{Cooperation with a guard}
\label{Figure:Example-3}
\end{figure}

The implementation was carried out with JAVA language. The figures \ref{Figure:Example-1}, \ref{Figure:Example-2} and \ref{Figure:Example-3} show three tested examples. The figure \ref{Figure:Example-1} gives an exemple for two robots with a simple cut and tasks parallelization. The figure \ref{Figure:Example-2} shows an exemple for two robots in the case of a complex cut leading to four connected components. Finally the figure \ref{Figure:Example-3} shows a deliberation example between a stuck robot and a guard.

\textsc{Note.}\ ---\ In these examples, the robots are represented by $\bullet$, the hached areas show the robots perception, the white areas show the decontaminated part of the environment whereas the black areas show the contaminated ones.

\section{Discussion and conclusion}
\label{Discussion}
In comparison to previous works such as \cite{LaValle-Lin-Guibas-Latombe-Motwani-97}, this cooperation protocol allows to minimize the number of robots by making them work as a team, to distribute and to share the exploration among all the robots. It is based on a ``least commitment'' strategy: indeed, extra robots are added if and only if cooperation fails and assistance is successively sought among the least constrained robots (idle, guard, explorer and then stuck robots). This leads to a better use of the robots resources, and an increase of performances and robustness due to the parallelization.\\

Furthermore, this cooperation protocol can be adapted for an unknown environment. In this case, the construction of the graphs can be done iteratively: each time new critical points are discovered during the exploration, they are added to the visibility and surveillance graphs respectively. The exploration and the construction of the visibility graph can be done simultaneously. When a robot reaches a delegation point, the team can still run a deliberation about the known environment. Further deliberations are undertaken as the known environment grows. Unlike \cite{Rajko-LaValle-00} where the algorithm provided works for only one robot and needs two successive steps (environment cartography and then search for an intruder), the simultaneous discovery and exploration of the environment can lead to decrease the covered distances.

As far as complexity is concerned, the construction of the graphs is $O(2^n)$ with $n$ the number of critical points of the environment. However, the complexity is tractable because practically the number of critical points remains small. This is a fortiori verified when exploring and constructing the graphs simultaneously. Another advantage of this distributed approach is that the calculation is shared between all the teammates. The protocol complexity (i.e. corresponding to the number of exchanged messages) is $O(n^2)$ in the worst case where $n$ is the number of robots.\\

Finally, the deliberation protocol can be adapted in order to take into account the robots specificities in terms of available resources. That is to say, their ability to answer to assistance requests does not only depend on their roles (guard, explorer, idle or stuck robot) but also on their available energy etc.

\newcommand{\etalchar}[1]{$^{#1}$}

\end{document}